\documentclass[runningheads]{llncs}
\usepackage{graphicx}

\usepackage{tikz}
\usepackage{comment}
\usepackage{amsmath,amssymb} 
\usepackage{color}
\usepackage{breakcites}
\usepackage[accsupp]{axessibility}  

\usepackage{subfigure}

\usepackage{booktabs}
\usepackage{multirow}

\newcommand{\Sref}[1]{Sec.~\ref{#1}}

\newcommand{\Fref}[1]{Fig.~\ref{#1}}
\newcommand{\Tref}[1]{Table~\ref{#1}}
\usepackage[pagebackref,breaklinks,colorlinks]{hyperref}

\def\mL{\mathcal{L}}

\def\1n{\mathbf{1}_n}
\def\0{\mathbf{0}}
\def\1{\mathbf{1}}

\newcommand{\cm}[1]{}

\newcommand{\mhoai}[1]{{\color{magenta}\textbf{[MH: #1]}}}

\newcommand{\myheading}[1]{\vspace{1ex}\noindent \textbf{#1}}

\newif\ifshowsolution
\showsolutiontrue

\ifshowsolution

\else

\fi

\usepackage[capitalize]{cleveref}
\crefname{section}{Sec.}{Secs.}
\Crefname{section}{Section}{Sections}
\Crefname{table}{Table}{Tables}
\crefname{table}{Tab.}{Tabs.}

\usepackage{adjustbox}

\begin{document}
\pagestyle{headings}
\mainmatter
\def\ECCVSubNumber{4169}  

\title{Exemplar Free Class Agnostic Counting} 

\titlerunning{Exemplar Free Counting}
%
\author{Viresh Ranjan\inst{1} \and
Minh Hoai\inst{1,2}}
\authorrunning{Ranjan et al.}
%
\institute{Stony Brook University,  USA \and
VinAI Research, Vietnam}


\maketitle
\begin{abstract}
We tackle the task of Class Agnostic Counting, which aims to count objects in a novel object category at test time without any access to labeled training data for that category. All previous class agnostic counting methods cannot work in a fully automated setting, and require computationally expensive test time adaptation. To address these challenges, we propose a visual counter which operates in a fully automated setting and does not require any test time adaptation. Our proposed approach first identifies exemplars from repeating objects in an image, and then counts the repeating objects. We propose a novel region proposal network for identifying the exemplars. After identifying the exemplars, we obtain the corresponding count by using a density estimation based Visual Counter.
We evaluate our proposed approach on FSC-147 dataset, and show that it achieves superior performance compared to the existing approaches. Our code and models will be made public.
\end{abstract}

\section{Introduction}
\label{sec:intro}
\begin{figure}[t]
    \centering
     \includegraphics[width=\linewidth]{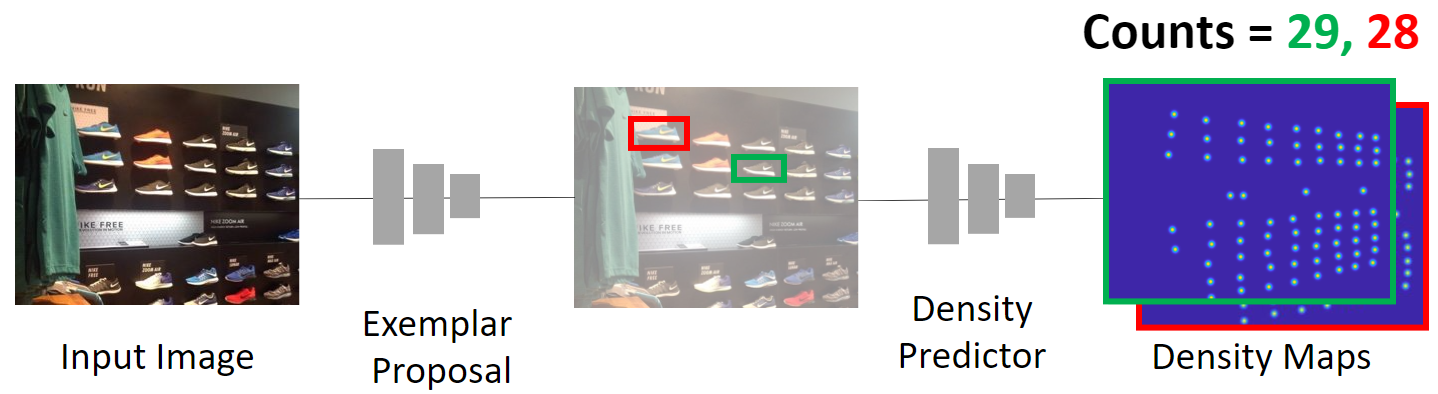}
     \vskip -0.1in
    \caption{\textbf{Exemplar Free Class Agnostic Counter}. Given an image containing instances of objects from unseen object categories, our proposed approach first generates exemplars from the repeating classes in the image using a novel region proposal network. Subsequently, a density predictor network predicts separate density maps for each of the exemplars. The total count for any exemplar, i.e. the number of times the object within the exemplar occurs in the image, is obtained by summing all the  values in the density map corresponding to that exemplar.
  \label{fig:motivation}
}
 \end{figure}

In recent years, visual counters have become more and more accurate at counting objects from specialized categories such as human crowd~\cite{Idrees_2013_CVPR,zhang2016single,idrees2018composition,ma2019bayesian}, cars~\cite{mundhenk2016large}, animals~\cite{arteta2016counting}, and cells~\cite{arteta2016detecting,xie2018microscopy,khan2016deep}. Most of these visual counters treat counting as a class-specific regression task, where a class-specific mapping is learned to map from an input image to the corresponding object density map, and the count is obtained by summing over the density map. However, this approach does not provide a scalable solution for counting objects from a large number of object categories because these visual counters can count only a single category at a time, and it also requires hundreds of thousands~\cite{zhang2016single} to millions of annotated training objects~\cite{wang2020nwpu,sindagi2020jhu} to achieve reasonably accurate performance for each category. A more scalable approach for counting objects from many categories is to use class-agnostic visual counters~\cite{lu2018class,ranjan2021learning}, which can count objects from many categories. 
But the downside of not having a predefined object category is that  these counters require a human user to specify what they want to count by providing several exemplars for the object category of interest. 
As a result, these class-agnostic visual counters cannot be used in any fully automated systems. Furthermore, these visual counters need to be adapted to each new visual category~\cite{lu2018class} or each test image~\cite{ranjan2021learning}, leading to  slower inference. 

In this paper, we present the first exemplar-free class-agnostic visual counter that is capable of counting objects from many categories, even for novel categories that have neither annotated objects at training time nor exemplar objects at testing time. Our visual counter does not require any human user in its counting process, and this will be very crucial for building fully automated systems in various applications in wildlife monitoring, healthcare and visual anomaly detection. For example, this visual counter can be used to alert environmentalists when a herd of animals with significant size pass by an area monitored by a wildlife camera. Another example is to use this visual counter to monitor for critical health conditions when any certain type of cells outgrows the other types. Unlike existing class-agnostic counters~\cite{lu2018class,ranjan2021learning}, our approach does not use any test time adaptation or finetuning.



At this point, a reader might wonder if it is possible to identify all possible exemplars in an image automatically by using a class-agnostic object detector such as a Region Proposal Network (RPN)~\cite{ren2015faster}, and run an existing class-agnostic visual counter using the detected exemplars to count all objects in all categories. Although this approach does not require a human's input during the counting process, it can be computationally expensive. This is because the RPNs usually produce a thousand or more of object proposals. And this in turn requires executing the class-agnostic visual counter at least a thousand times, a time-consuming and computationally demanding process.

To avoid this expensive procedure, we develop in this paper a novel convolutional network architecture called \textbf{Rep}etitive \textbf{R}egion \textbf{P}roposal \textbf{N}etwork (RepRPN), which can be used to automatically identify few exemplars from the most frequent classes in the image. RepRPN is used at the first stage of our proposed two-stage visual counting algorithm named RepRPN-Counter. We use a density estimation based Visual Counter as the second stage of the RepRPN-Counter, which predicts a separate high resolution density map for each exemplar. Given an input image, RepRPN considers multiple region proposals, and compute the objectness and repetition scores for each  proposal. The repetition score of a proposal is defined as the number of times the object contained within the proposal occurs in the image. The proposals with the highest repetition scores are chosen as the exemplars, and the second stage density predictor estimates the density maps only for the chosen exemplars with high repetition scores. This exemplar generation procedure relies on the underlying assumption that in an image containing different classes with varying counts, the classes of interest are the ones having larger counts. Compared to the traditional RPN~\cite{ren2015faster}, RepRPN is better suited for visual counting task, since it can significantly reduce the training and inference time for any two-stage counter. Furthermore, RepRPN can serve as a fast visual counter for applications which can tolerate some margin of error and do not require the localization information conveyed by density maps. Note that the second stage predictor of our visual counter estimates a separate density map for each of the chosen exemplars.

While training RepRPN-Counter, another technical challenge that we need to overcome is the lack of proper annotated data. The only dataset suitable for training class-agnostic visual counters is FSC-147~\cite{ranjan2021learning}, which contains annotation for a single object category in each image, and may contain unannotated objects from other categories. To obtain annotation for unannotated objects in the FSC-147 dataset, we propose a novel knowledge transfer strategy where we use a RepRPN trained on a large scale object detection dataset~\cite{lin2014microsoft} and a density prediction network~\cite{ranjan2021learning} trained on FSC-147 as teacher networks.

In short, the contributions of this paper are threefold: (1) we develop the first exemplar free class agnostic visual counter for novel categories that have neither annotated objects at training time nor exemplar objects at testing time; (2) we develop a novel architecture to simultaneously estimate the objectness and repetition scores of each proposal; (3) we propose a knowledge transfer strategy to handle unannotated objects in the FSC-147 dataset.

\section{Related Work}

\myheading{Visual Counting.} Most previous methods for visual counting focus on specific categories~\cite{zhang2019attentional,wan2019adaptive,m_Ranjan-etal-ACCV20,m_Abousamra-etal-AAAI21,ma2019bayesian,zhang2016single,ranjan2018iterative,babu2018divide,sam2017switching,li2018csrnet,liu2018leveraging,cao2018scale,ranjan2019crowd,shi2019revisiting,liu2019context,wang2019learning,song2021rethinking,wan2021generalized,wang2021uniformity}. These visual counters can count a single category at a time, and require training data with hundreds of thousands~\cite{zhang2016single} to millions of annotated instances~\cite{wang2020nwpu} for every visual category, which are expensive to collect. These visual counters cannot generalize to new categories at test time, and hence, cannot handle our class agnostic counting task. To reduce the expensive annotation cost, some of these methods focus on designing unsupervised~\cite{liu2018leveraging} and semi-supervised tasks~\cite{liu2020semi} for visual counting. However, these methods still require a significant amount of annotations and training time for each new category.

\myheading{Class Agnostic Counting.} Most related to ours is the previous works on class agnostic counting~\cite{lu2018class,ranjan2021learning}, which build counters that can be trained to count novel classes using relatively small number of examples from the novel classes. Lu and Zisserman~\cite{lu2018class} proposed a Generic Matching Network (GMN) for class-agnostic counting, which follows a two-stage training framework where the first stage is trained on a large-scale video object tracking data, and the second stage consists of adapting GMN to a novel object class. GMN uses labeled data from the novel object class during the second stage, and only works well if several dozens to hundreds of examples are available for the adaptation. 
Few-shot Adaptation and Matching Network (FamNet)~\cite{ranjan2021learning} is a recently proposed class agnostic few-shot visual counter which generalizes to a novel category at test time given only a few exemplars from the category. However, FamNet is an interactive visual counter which requires an user to provide the exemplars from the test image. Both GMN and FamNet require test time adaptation for each new class or test image, leading to slower counting procedures. 

\myheading{Zero-Shot Object Detection.} Also related to ours is the previous work on zero-shot object detection
~\cite{bansal2018zero,zhu2019zero,rahman2018zero}. Most of these approaches~\cite{bansal2018zero,rahman2018zero} use a region proposal network to generate class-agnostic proposals, and map the features from the proposals to a semantic space where they can be directly compared with semantic word embeddings of novel object classes. However, all of these zero-shot detection approaches require access to the semantic word embeddings for the test classes, and cannot work for our class agnostic counting task where the test classes are not known a priori. 

\myheading{Few-Shot Learning.} Also related to ours is the previous works on few-shot learning~\cite{lake2015human,koch2015siamese,santoro2016one,finn2017model,ravi2016optimization}, which aim to adapt classifiers to novel categories based on only a few labeled examples.
One of the meta learning based few-shot approaches, Model Agnostic Meta Learning (MAML)~\cite{finn2017model}, has been adapted for class-agnostic counting~\cite{ranjan2021learning}. MAML focuses on learning parameters which can adapt to novel classes at test time by doing only a few gradient descent steps. Although these few-shot methods reduce the labeled data needed to generalize to new domains, most of these approaches cannot be used for our class agnostic counting task due to the unavailability of labeled data from the novel test class.
\section{Proposed Approach}

\begin{figure*}[!t]
    \includegraphics[width=\textwidth]{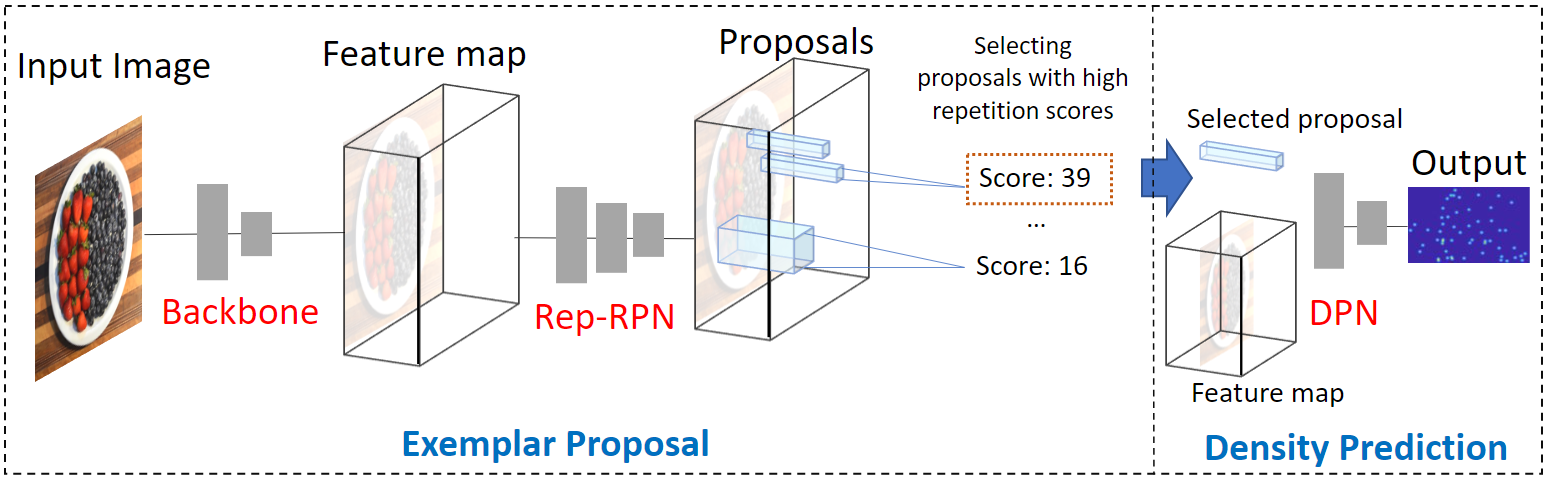}
    \vskip -0.1in
    \caption{{\bf RepRPN-Counter} is a two-stage Exemplar Free Class Agnostic Counter. RepRPN-Counter has two key components: 1) Repetitive Region Proposal Network (RepRPN) and 2) Density Prediction Network (DPN). RepRPN predicts repetition score and objectness score for every proposal. Repetition score is used to select few proposals, called exemplars, from the repeating classes in the image. The DPN predicts a separate density map for any proposal selected by the RepRPN. The total count for any proposal, i.e. the number of times the object within the proposal occurs in the image, is obtained by summing all the  values in the density map corresponding to that proposal. The DPN ignores proposals which are less likely to contain repetitive objects, so as to reduce the time required for training and evaluation. To keep things simple, we have shown the density prediction step for a single proposal. In reality, several density maps are predicted by the DPN, one for every selected proposal. 
  \label{fig:RepRPNBlock}}
 \end{figure*}  


We propose an exemplar-free class-agnostic visual counter called RepRPN-Counter. Given an image containing one or more repetitive object categories, RepRPN-Counter predicts a separate density map for each of the repetitive categories. The object count for the repetitive categories can be obtained by simply summing up the corresponding density map. For a category that is counted, RepRPN-Counter also provides the bounding box for an example from the category. 


RepRPN-Counter consists of two key components: 1) a Repetitive Region Proposal Network (RepRPN) for identifying exemplars from repetitive objects in an image, along with their approximate count; and 2) a Density Prediction Network (DPN) that predicts a density map corresponding to any exemplar produced by the RepRPN. 

For the rest of this  section, we will describe the architecture of RepRPN in \Sref{reprpn}, the architecture of RepRPN-Counter in \Sref{reprpn-counter}, the knowledge transfer approach for handling incomplete annotation in \Sref{missinglabels}, and the overall training strategy in \Sref{training}

\begin{figure}[!tb]
\hfill 
   \includegraphics[height=2cm,width=0.16\linewidth]{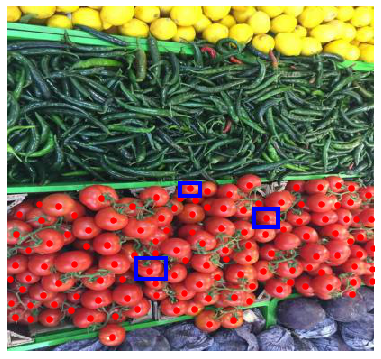}
  \hfill
   \includegraphics[height=2cm,width=0.24\linewidth]{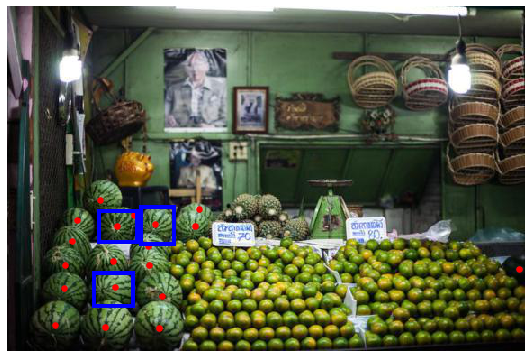}
  \hfill
    \includegraphics[height=2cm,width=0.20\linewidth]{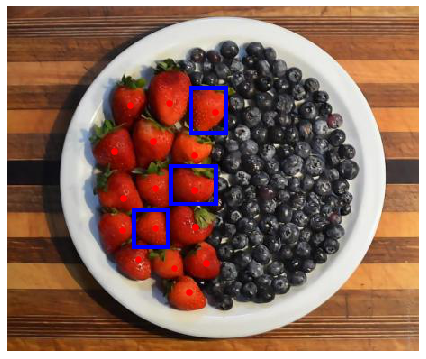}
  \hfill
\includegraphics[height=2cm,width=0.22\linewidth]{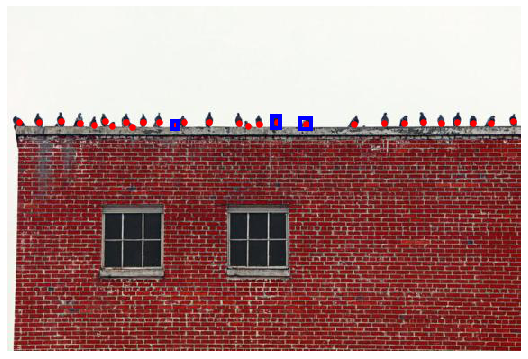} \hfill \hfill 
\vskip -0.1in 
 \caption{{\bf Missing labels in the FSC-147 dataset}. Each image in the dataset comes with bounding box annotations for the exemplar objects(shown in blue), and dot annotations for all objects belonging to the same category as the exemplar. For each image, objects of only a single class are annotated. We present a knowledge transfer strategy to deal with incomplete annotation.   \label{fig:annotation_examples} }
\end{figure}

\subsection{Repetitive Region Proposal Networks  \label{reprpn}}
 Repetitive Region Proposal Network (RepRPN) proposes exemplars from the repetitive object classes in an image. RepRPN takes as input convolutional feature representation of an image computed by the Resnet-50 backbone~\cite{He-et-al-CVPR16}, and predicts proposal bounding boxes along with objectness and repetition scores for every proposal at every anchor location. The objectness score is the probability of the proposal belonging to any object class and not the background class. The repetition score refers to the number of times the object within the proposal occurs in the image. For example, consider an image with~$m$ cats and~$n$ oranges. The RepRPN should predict $m$ as the repetition score for any cat proposal, and~$n$ as the repetition score for any orange proposal. The repetition score is used to select exemplars from the repetitive classes in the image, i.e. the proposals with the highest repetition score are chosen as the exemplars. The original RPN formulation~\cite{ren2015faster} uses a fixed window around an anchor location to predict the proposal boxes and objectness scores. However, this fixed sized window does not cover the entire image, and it does not contain sufficient information to predict the repetition score. This has been verified in our experiment where an RPN using only fixed-size window over convolutional features was unable to predict the repetition score. Predicting repetition score would require access to information from the entire image. To obtain this global information efficiently, we make use of the Encoder Self-Attention layers~\cite{vaswani2017attention}. Given a feature vector at any location in the convolutional feature map, self-attention layers can pool information from \textit{similar} vectors from the entire image, and can be used to estimate repetition score at any anchor location. To apply self-attention, we first transform the convolutional features into a sequence of length $n$: $S \in R^{n \times d}$. To preserve positional information, we concatenate  appropriate $\frac{d}{2}$-dimensional row and column embeddings, resulting in $d$ dimensional positional embeddings which are added with the sequence $S$. We refer to the resulting embeddings as $X \in R^{n\times d}$. 

Given the sequence $X$, the self-attention layer first transforms $X$ into query ($X_Q$), key ($X_K$), and value ($X_V$) matrices by multiplying $X$ with matrices $W_Q$, $W_K$, and $W_V$:
\begin{align}\label{eqn:eqn1}
 X_Q = X W_Q,\quad X_K = X W_K, \quad X_V = X W_V.
\end{align}

The self-attention layer outputs a new sequence $U$ where the $i^{th}$ element in the output sequence is obtained as a weighted average of the value sequence, and the weights are decided based on the similarity between the $i^{th}$ query element and the key sequence.
The output sequence $U$ is computed efficiently by doing two matrix multiplications:
\begin{equation}\label{eqn:eqn2}
U = softmax(X_Q X_K^T)X_V.   
\end{equation}
Tensor $U$ will be reshaped into a tensor $U'$ that has the same spatial dimensions as the input convolutional feature map. Tensor $U'$ will be forwarded to the bounding box regression, objectness prediction, and repetition score prediction heads. Each prediction head consists of a single $1{\times}1$ convolutional layer. At each anchor location in the image, we consider $k$ anchors boxes. For each anchor box, we will output an objectness score, a repetition score, and bounding box coordinates. The repetition score is used to identify proposals containing the repetitive objects in the image, i.e. proposals with a large repetition score contain repetitive objects.

\subsection{RepRPN-Counter}\label{reprpn-counter}

As shown in \Fref{fig:RepRPNBlock}, RepRPN-Counter consists of a Resnet-50 feature backbone, a RepRPN proposal network, and a Density Prediction Network (DPN). The RepRPN and the DPN share the same feature backbone. The RepRPN  provides the DPN with the bounding box locations of the proposals with large repetition scores, also called exemplars, and the DPN predicts a separate density map for each exemplar. DPN is trained and evaluated on only the chosen exemplars, and not all the proposals, so as to reduce the training and inference time. Similar to the previous works on class-agnostic counting~\cite{lu2018class,ranjan2021learning}, DPN combines the convolutional features of an exemplar, with the convolutional features of the entire image to predict the density map for the exemplar. The exemplar features are obtained by performing ROI pooling on the convolutional features computed by the backbone, at the locations defined by the exemplar bounding boxes. The exemplar features are correlated with the image features, and the resulting correlation map is propagated through the DPN. The DPN is a fully convolutional network consisting of five convolutional layers and three upsampling layers (more architecture details are provided in the Supplementary submission), and the predicted density maps have the same spatial dimensions as the the input image. Note that the DPN predicts several density maps, one for each exemplar. The overall count for an object class pertaining to an exemplar is obtained by simply summing all the values in the density map corresponding to the exemplar. The DPN is not evaluated on the proposals with a low repetition score. For such proposals, the repetition score can be used as the final count.

\subsection{Knowledge transfer for handling missing labels}\label{missinglabels}

The only existing dataset consisting of images of densely populated objects from many visual categories  that can be used for training class agnostic visual counters is FSC-147~\cite{ranjan2021learning}. However, it is not trivial to train RepRPN-Counter on FSC-147 because of the missing labels in the dataset. FSC-147 comes with two types of annotations for each image: a few exemplar bounding boxes to specify the object category to be counted, and dot annotations for all of the objects belonging to the same category as the specified exemplars. However, an image may contain objects from another category that has not been annotated, as shown in \Fref{fig:annotation_examples}. Given the missing labels, forcing RepRPN-Counter to predict zero count for the unannotated objects may degrade the performance of the counter.

We use knowledge transfer from teacher networks to address the incomplete annotation issue. We first train a RepRPN on the MSCOCO object detection  dataset~\cite{lin2014microsoft}. The MSCOCO training set consists of over 82K natural images from 80 visual categories, and the RPNs trained on this large dataset have been shown to generalize to previously unseen classes, thereby proving useful for tasks like zero-shot object detection~\cite{rahman2018zero}. We use the RepRPN trained on MSCOCO as a teacher network for generating the target labels for the objectness scores and the repetition scores for those proposals not intersecting with the annotated objects in the FSC-147 dataset. 
To get the target density maps corresponding to the unannotated proposals, we use the pretrained class-agnostic visual counter FamNet~\cite{ranjan2021learning}, which can predict the density map for a novel object class given only a single exemplar. When needed, an unannotated proposal is fed into FamNet, and the output of FamNet is used as the target density map for training the proposed network RepRPN-Counter. 


\subsection{Training Objective}\label{training}
RepRPN-Counter is trained in two stages. The first stage consists of training the RepRPN. Once trained, the RepRPN is kept frozen and used to generate exemplars for the density estimation network DPN. The second stage of training consists of training the DPN to predict the density map for every exemplar.

\myheading{Training objective for RepRPN.} For the $i^{th}$ anchor box, the outputs of the RepRPN are the objectness score $y_i$, the bounding box coordinates $b_i$, and the repetition score $c_i$. Let the corresponding ground truth labels be $y_i^*$, $b_i^*$, $c_i^*$. We follow the same protocol as used in Faster RCNN~\cite{ren2015faster} for obtaining the binary objectness label $y_i^*$, and the same parameterization for the bounding box coordinates $b_i$. $c_i^*$ is the number of times the object within the anchor box, if any, occurs in the image. Since predicting $c_i$ requires access to global information about the image, RepRPN makes use of self-attentional features as described in \Sref{reprpn}. The training loss for the $i^{th}$ anchor box is: 
\begin{equation}
\mL_{RepRPN} =
\lambda \mL_{cls}(y_i,y_i^*) + \lambda \mL_{reg}(b_i,b_i^*) + \mL_{reg}(c_i,c_i^*),
  \label{rpn}
  \end{equation}
where $\mL_{cls}$ is the binary cross entropy loss, and $\mL_{reg}$ is the smooth $L_1$ loss. When training RepRPN on the FSC-147 dataset, the labels $y_i^*$ and $c_i^*$ for the positive anchors are obtained using the ground-truth annotation of FSC-147. Note that in the FSC-147 dataset, only three exemplars per image are annotated with bounding boxes, while the rest of the objects are annotated with a dot around their center. We obtain the bounding boxes for all the dot annotated objects by placing a bounding box of the average exemplar size around each of the dots. For anchors not intersecting with any of the annotated bounding boxes in FSC-147, $y_i^*$ and $c_i^*$ labels are obtained using a teacher RepRPN, which has been pre-trained on the MSCOCO dataset~\cite{lin2014microsoft}. 
  
  \myheading{Training objective for DPN.}
  Given an exemplar bounding box $b_i$ and the feature map $U$ for an input image $I$ of size $H{\times}W$, the density prediction network DPN predicts a density map 
$Z_{b_i} = f(U,b_i)$ of size $H{\times}W$. The training objective for the DPN is based on the mean square error: 
\begin{align}
    \mL_{mse}(Z_{b_i},Z^*) = \frac{1}{HW} \sum_{r=1}^H\sum_{c=1}^{W} ( Z_{b_i}(r,c) - Z^{*}(r,c))^2, \nonumber 
\end{align}
where $Z^{*}$ is the target density map corresponding to $Z_{b_i}$. If the exemplar $b_i$ intersects with any annotated object, $Z^{*}$ is obtained by convolving a Gaussian kernel with the corresponding dot annotation map. Note that Gaussian blurred dot annotation maps are commonly used for training density estimation based visual counters~\cite{zhang2016single,idrees2018composition,ranjan2018iterative,lu2018class,Liu-etal-CVPR18}. For cases where $b_i$ does not intersect with any annotated object, we use the pretrained FamNet~\cite{ranjan2021learning} as a teacher network for obtaining $Z^{*}$. The FamNet teacher can predict a density map, given an exemplar $b_i$ and an input image $I$.

\subsection{Implementation details}
 For training, we use Adam optimizer~\cite{kingma2014adam} with a learning rate of $10^{-5}$ and batch size of one. We use the first four convolutional blocks from the ImageNet pre-trained ResNet-50~\cite{He-et-al-CVPR16} as the backbone. We keep the backbone frozen during training, since finetuning the backbone would yield poor results. This is because the backbone has feature maps suitable for detecting a large  number of classes, and finetuning the backbone leads to specialization towards FSC-147 training classes, resulting in poor performance on the novel test classes. 
 
 The weights of the RepRPN and DPN are initialized from a zero mean univariate Gaussian with standard deviation of $10^{-3}$. RepRPN uses five self-attention transformer layers, each with eight heads. For training the RepRPN, we use four anchors sizes of $32,64,128,256$ and two aspect ratios of $0.5, 1, 2$. We sample a batch of 96 anchors from each image during training. Training is done for 1000 epochs.

\section{Experiments}

\subsection{Dataset}
We perform experiments on the recently proposed FSC147 dataset~\cite{ranjan2021learning}, which was originally proposed for the exemplar based class-agnostic counting task. The FSC147 dataset consists of 6135 images from 147 visual categories, which are split into train, val, and test splits comprising of 89, 29, and 29 classes respectively. There are no common categories between the train, val, and test sets. The mean and maximum counts for images in the dataset are 56 and 3701, respectively. We train our model on the train set, and evaluate it on the test and val sets. Each image comes with annotations for a single object category of interest only, which consists of several exemplar bounding boxes and complete dot annotation for the objects of interest in the image.  Since our goal is to build an exemplar free counter, unlike previous methods~\cite{lu2018class,ranjan2021learning}, we do not use human annotated exemplars as an input to our counter. 

\subsection{Evaluation Metrics} 

We use the Top-$k$ version of  Mean Absolute Error (MAE) and Root Mean Squared Error (RMSE) to compare the performance of the different visual counters. MAE and RMSE are defined as follows.
$MAE = \frac{1}{n}\sum_{i=1}^{n} \lvert y_i - \hat{y}_i \rvert; 
RMSE = \sqrt[]{\frac{1}{n}\sum_{i=1}^{n} (y_i - \hat{y}_i)^2}, 
$
where $n$ is the number of test images, and $y_i$ and $\hat{y}_i$ are the ground truth and predicted counts. MAE and RMSE
are the most commonly used metrics for counting task~\cite{zhang2016single,ma2019bayesian,lu2018class,ranjan2021learning}. However, RepRPN-Counter predicts several density maps and corresponding  counts, one for each selected proposal. Given $k$ predicted counts from $k$ proposals, we compute Top-$k$ MAE and RMSE by first selecting those proposals from the top $k$ proposals which have an IoU ratio of at least 0.3 with any ground truth boxes, and average the counts corresponding to the selected proposals to get the predicted count~$\hat{y}_i$. In case none of the k proposals intersect with any ground truth boxes, we simply average all of the k counts to get~$\hat{y}_i$. 
\begin{table*}[t]
\caption{{\bf Comparing RepRPN-Counter to class-agnostic counters.} FamNet, GMN and MAML are exemplar based class-agnostic counters which have been adapted and trained for the exemplar-free setting, where a RPN is used for generating exemplars. We report the Top-1, Top-3 and Top-5 MAE and RMSE metrics on the val and test sets of FSC-147 dataset. RepRPN-Counter consistently outperforms the competing approaches.
\label{tab:main1}}
\vskip -0.05in
\begin{adjustbox}{width=\linewidth,center}
\begin{tabular}{lrrr|rrr|rrr|rrr}
\toprule
\multirow{2}{*}{Method} & \multicolumn{3}{c}{MAE (Val set)} & \multicolumn{3}{c}{RMSE (Val set)} & \multicolumn{3}{c}{MAE (Test set)} & \multicolumn{3}{c}{RMSE (Test Set)} \\
                        & Top1     & Top3     & Top5     & Top1      & Top3     & Top5     & Top1      & Top3     & Top5     & Top1      & Top3      & Top5     \\
\midrule
GMN &43.25 &40.96 &39.02 & 114.52&108.47 & 106.06&43.35 &39.72 & 37.86& 145.34& 142.81& 141.39
          \\

MAML  & 34.96 & 33.16 & 32.44 &  \textbf{98.83}& 101.80& 101.08&37.38 & 33.27& 31.47&133.89 & 131.00&129.31
          \\
 FamNet(pretrained) & 47.66& 42.85& 39.52& 125.54 & 121.59& 116.08 &50.89 & 42.70& 39.38& 150.52& 146.08& 143.51
          \\
FamNet & 34.51 & 33.17& 32.15& 99.87& 99.31& 98.75&  35.81 & 33.32& 32.27& 133.57& 132.52&131.46         \\

  RepRPN-Counter & \textbf{31.69} &  \textbf{30.40} &  \textbf{29.24} & 100.31 &  \textbf{98.73} &  \textbf{98.11} &  \textbf{28.32} &  \textbf{27.45} &  \textbf{26.66} &  \textbf{128.76} &  \textbf{129.69} &  \textbf{129.11} \\
\bottomrule
\end{tabular}
\end{adjustbox}

\end{table*}

\subsection{Comparison with class-agnostic visual counters}\label{expClassAgnostic}
We compare our proposed RepRPN-Counter with the previous class-agnostic counting methods~\cite{lu2018class,ranjan2021learning,finn2017model} on the task of counting objects from novel classes. We do not compare with class-specific counters~\cite{zhang2016single,ma2019bayesian} because such counters cannot handle novel classes at test time. Furthermore, these counters require hundreds~\cite{zhang2016single} or thousands~\cite{wang2020nwpu,sindagi2020jhu} of images per category during training, while FSC-147 dataset contains an average of only 41 images per category. 

GMN~\cite{lu2018class}, FamNet~\cite{ranjan2021learning}, and MAML~\cite{finn2017model,ranjan2021learning} are exemplar based counters which can predict density map for any unseen object category based on few exemplars of the object category from the same image. These counters were originally proposed to work with human provided exemplars as an input to the counter. In order to make these exemplar based counters work in our exemplar free setup, we modify GMN, FamNet, and MAML based visual counters by replacing human provided exemplars with RPN~\cite{ren2015faster} generated exemplars. We use the RPN of Faster RCNN~\cite{ren2015faster} to generate the proposals for the competing approaches, and use the top $k$ proposals with the highest objectness score as the exemplars. For fair comparison, both the RPNs used with the competing approaches as well as the RepRPN are pre-trained on the MSCOCO dataset~\cite{lin2014microsoft}. We do not use MSCOCO to train the DPN. We train the competing approaches and our proposed approach on the train set of FSC-147, and report the results on the val and test sets of FSC-147. We also compare our method with a pre-trained version of FamNet originally trained on the few-shot counting task. Following~\cite{ranjan2021learning}, all of the methods are trained with three proposals, and evaluated with 1, 3, and 5 proposals. We report the Top-$1$, Top-$3$ and Top-$5$ MAE and RMSE values in \Tref{tab:main1}. 

As can be seen from \Tref{tab:main1}, our method RepRPN-Counter outperforms all of the competing methods. The pre-trained FamNet performs the worst, even though it was trained on the same FSC-147 training set. This shows that simply combining pre-trained exemplar based class agnostic counters with RPN-based exemplars does not provide a reasonable solution for the exemplar-free setting. When retrained specifically for the exemplar-free setting, the performance of FamNet significantly improves when compared to its pre-trained version. GMN performs worse than the other baselines, possibly due to the need for more examples for the adaptation process. This observation was earlier reported for exemplar based class-agnostic counting task as well~\cite{ranjan2021learning}.

\setlength{\tabcolsep}{10pt}
\begin{table}[t]
\caption{{\bf Comparing RepRPN-Counter with pre-trained object detectors}, on Val-COCO and Test-COCO subsets of \mbox{FSC-147}, which only contain COCO classes. Pre-trained object detectors are available for these COCO classes. For RepRPN-Counter, we use the density map corresponding to the proposal with the highest repetition score. Without access to any labeled data from these COCO classes, our proposed approach outperforms all of the object detectors which are trained using the entire COCO train set containing a large number of images from these COCO classes
\label{tab:detectors}}
\vskip -0.1in
\centering
\begin{tabular}{lcccc}
\toprule
          &  \multicolumn{2}{c}{ Val-COCO Set} &  \multicolumn{2}{c}{ Test-COCO Set} \\
         \cmidrule(lr){2-3} \cmidrule(lr){4-5} 
    Method &  MAE          & RMSE    & MAE          & RMSE         \\

\midrule 
Faster R-CNN & 52.79 & 172.46 & 36.20 & 79.59 \\
RetinaNet & 63.57 & 174.36 & 52.67 & 85.86 \\
Mask R-CNN & 52.51 & 172.21 & 35.56 & 80.00 \\
Detr &  58.35 & 175.97 & 45.51 & 96.57 \\
RepRPN-Counter (Ours)  & \textbf{50.72} & \textbf{160.95} & \textbf{25.29} & \textbf{56.98}  \\

\bottomrule 
\end{tabular}

\end{table}

\subsection{Comparison with object detectors}\label{expDetectors}

One approach to counting is to use a detector and count the number of detections in an image. However, it requires thousands of examples to train an object detector, and the detector-based counters cannot be used for novel object classes. That being said, we compare RepRPN-Counter with object detectors on a subset of COCO categories from the validation and test sets of FSC-147. These subsets are called Val-COCO and Test-COCO, containing 277 and 282 images respectively. We compare our approach with the official implementations~\cite{wu2019detectron2} of MaskRCNN~\cite{He-etal-ICCV17}, FasterRCNN~\cite{Ren-etal-NIPS15}, RetinaNet~\cite{lin2017focal}, and Detr~\cite{detr}. The results are shown in \Tref{tab:detectors}. Without any access to labeled data from these COCO classes, our proposed method still outperforms the object detectors that have been trained using the entire COCO train set containing thousands of images from these COCO classes. Detr~\cite{detr} performs worse than some of the earlier object detectors because Detr uses a fixed number of query slots (usually 100), which limits the maximum number of objects it can detect, while FSC-147 has images containing thousands of objects.

\subsection{Comparing RepRPN with RPN}
We are also interested in checking if RepRPN can boost the performance of class-agnostic visual counters other than RepRPN-Counter. For this experiment, we replace RPN~\cite{ren2015faster} with RepRPN for GMN, FamNet, and MAML, and report the Top-1 MAE and RMSE scores on the FSC-147 test set. The results are presented in \Tref{tab:RepRPN}. Using RepRPN instead of RPN leads to significant boost in the performance for all  class-agnostic visual counters. This suggests that RepRPN is much better suited for the exemplar proposal for exemplar free counting task in comparison to RPN. Also, RepRPN works well with different types of class agnostic counters, including the proposed RepRPN-Counter, GMN, FamNet, and MAML. 
\setlength{\tabcolsep}{10pt}
\begin{table}[!t]
\caption{{\bf Comparing RepRPN with RPN}, on the test set of FSC-147. Using RepRPN instead of RPN leads to significant boost in performance
\label{tab:RepRPN}}
\vskip -0.1in
\centering
\begin{tabular}{lcccc}
\toprule
          &  \multicolumn{2}{c}{RPN} &  \multicolumn{2}{c}{RepRPN} \\
         \cmidrule(lr){2-3} \cmidrule(lr){4-5} 
    Method &  MAE          & RMSE    & MAE          & RMSE         \\

\midrule 
GMN & 43.35 & 145.34 & 32.17 & 137.29 \\
MAML & 37.38 & 133.89 & 32.09 & 141.03 \\
FamNet (pre-trained) & 50.89 & 150.52 & 38.64 & 144.27 \\
FamNet & 35.81 & 133.57 & 32.94 & 132.82 \\
\bottomrule 
\end{tabular}
\end{table}

\subsection{Ablation Studies}
Our proposed RepRPN-Counter consists of two primary components: the RepRPN for exemplar proposal and the DPN for density prediction. Furthermore, our proposed knowledge transfer approach allows us to deal with unannotated objects in the FSC-147 dataset. In \Tref{tab:ablationA}, we analyze the contribution of these components on the overall performance. The RepRPN baseline uses the repetition score as the final count. We propose to use RepRPN with DPN, but one can replace RepRPN by RPN~\cite{ren2015faster} to get the method RPN+DPN. One can assume there are no unannotated objects in the FSC-147 dataset, and train our proposed RepRPN-Counter on FSC-147 without any knowledge transfer.
As can be seen from \Tref{tab:ablationA}, all the components of RepRPN-Counter are useful, and the best results are obtained when all the components are present. RPN+DPN performs much worse than RepRPN+DPN, which shows that RepRPN is better suited for our counting task than RPN.
\setlength{\tabcolsep}{3pt}
\begin{table}[t]
\caption{{\bf Analyzing individual components of RepRPN-Counter on the overall performance on the test set of FSC-147.} RPN + DPN refers to the case where we replace RepRPN from our proposed approach with the RPN from Faster RCNN. As can be seen, RepRPN is a critical component of our proposed approach, and replacing it with RPN decreases the performance significantly. RepRPN+DPN-NoKT refers to the method when we do not use any knowledge transfer, which leads to a drop in performance. This shows the usefulness of the proposed knowledge transfer strategy.\label{tab:ablationA}}
\vskip -0.1in
\begin{adjustbox}{width=\columnwidth,center}
\begin{tabular}{lllllll}
\toprule
\multirow{2}{*}{Method} & \multicolumn{3}{c}{MAE} & \multicolumn{3}{c}{RMSE} \\
 \cmidrule(lr){2-4}
  \cmidrule(lr){5-7}
                        & Top1  & Top3  & Top5 & Top1  & Top3  & Top5  \\
\midrule
RepRPN+DPN (proposed) & 28.32  & 27.45  & 26.66 & 128.76 & 129.69 & 129.11 \\
RepRPN+DPN-NoKT (no knowledge transfer) & 29.52  & 28.80  & 28.42 & 132.76 & 131.03 & 130.82 \\
RPN+DPN                     & 35.81  & 33.32  & 32.27 & 133.57 & 132.52 & 131.46 \\
RepRPN  (without DPN)                & 29.60  & 29.18  & 28.95 & 136.25 & 136.21 & 136.26 \\
\bottomrule
\end{tabular}
\end{adjustbox}

\end{table}

\newcommand\qualextwo{0.23\textwidth}
\newcommand\qualexthree{0.15\textwidth}
\newcommand\heightqualexa{2.6cm}

\begin{figure}[!htb]  
\centering
\makebox[\qualextwo]{Image}
\makebox[\qualextwo]{Prediction}
\makebox[\qualextwo]{Image}
\makebox[\qualextwo]{Prediction}

    \includegraphics[width=\qualextwo,height=\heightqualexa]{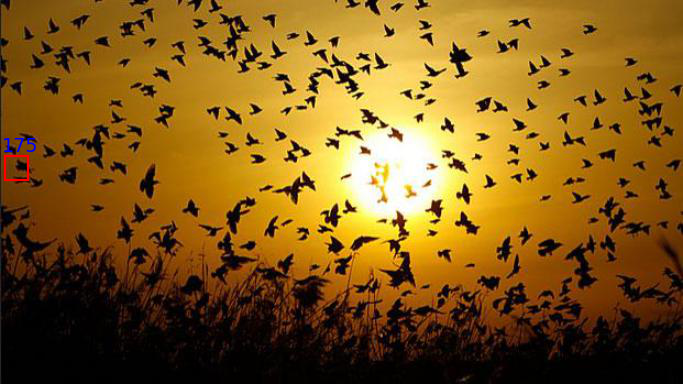}
    \includegraphics[width=\qualextwo,height=\heightqualexa]{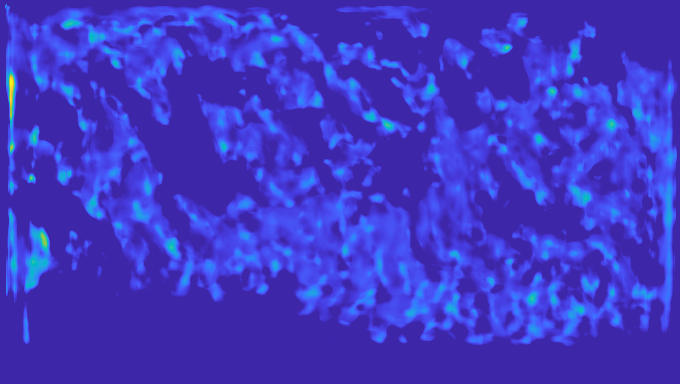}     \includegraphics[width=\qualextwo,height=\heightqualexa]{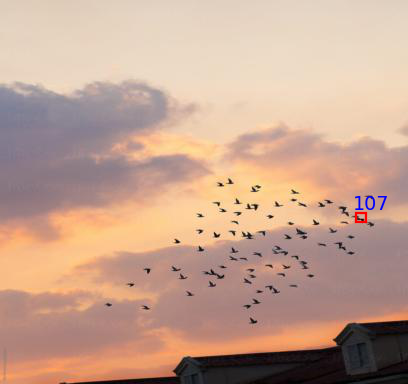}
    \includegraphics[width=\qualextwo,height=\heightqualexa]{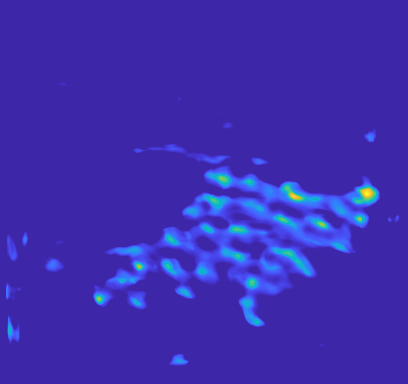} \\    
\makebox[\qualextwo]{GT Count: 298}
\makebox[\qualextwo]{Rep: 175, DPN: 331}
\makebox[\qualextwo]{GT Count: 83}
\makebox[\qualextwo]{Rep: 107, DPN: 83} \\ \vspace{2ex}

    \includegraphics[width=\qualextwo,height=\heightqualexa]{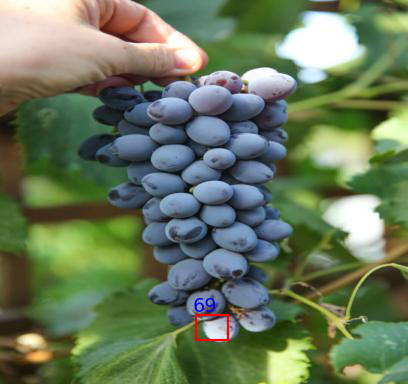}
    \includegraphics[width=\qualextwo,height=\heightqualexa]{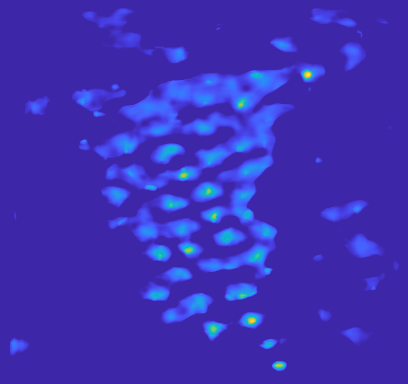}
    \includegraphics[width=\qualextwo,height=\heightqualexa]{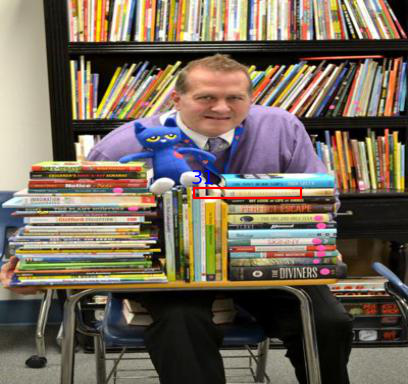}
    \includegraphics[width=\qualextwo,height=\heightqualexa]{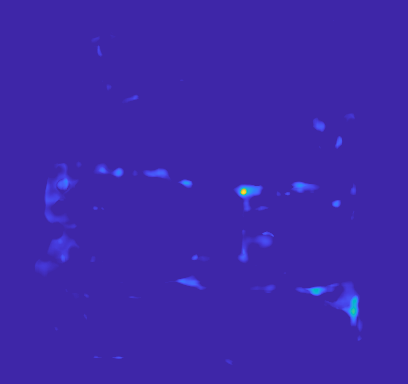} \\    
\makebox[\qualextwo]{GT Count: 55}
\makebox[\qualextwo]{Rep: 69, DPN: 61}
\makebox[\qualextwo]{GT Count: 206}
\makebox[\qualextwo]{Rep: 31, DPN: 10} \\ 

  \vskip -0.1in
  \caption{{\bf Input images and the density maps predicted by RepRPN-Counter.} Also shown in red  are the selected proposals. Rep is the repetition score predicted by RepRPN, while DPN is the count obtained by summing the final density map.}
  \label{fig:QualitativeDPN}
\end{figure}


\subsection{Qualitative Results}

In \Fref{fig:QualitativeDPN}, we present a few input images, the proposal with the highest repetition score generated by RepRPN for each image, and the corresponding density map generated by the density prediction network. RepRPN-Counter performs well on the first three test cases. But it fails on the last one, because the aspect ratio of the chosen proposal is very different from the majority of the objects of interest.

In \Fref{fig:QualitativeMore}, we show the RepRPN proposal with the highest repetition score for several images from the Val and Test set of FSC-147. We also show the repetition score and the corresponding ground truth count. First three rows contain test cases where the repetition score is close to the groundtruth count. The last row shows test cases which proved to be harder for RepRPN. RepRPN does not perform well in some cases when the objects are extremely small in size. Since RepRPN, and RPN in general, uses a fixed set of anchor sizes and aspect ratios, they may fail at detecting extremely small objects. It is also difficult for RepRPN to handle extreme variation in scale within the image, as evident from the first two failure cases of the last row.

\newcommand\heightqualex{2.5cm}

\begin{figure*}[!th]  
\centering

    \includegraphics[width=\qualextwo,height=\heightqualex]{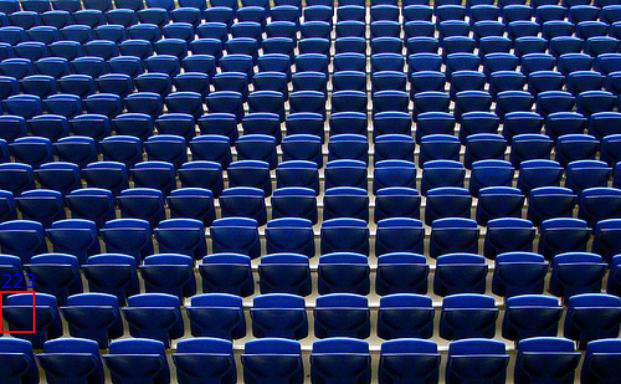}
    \includegraphics[width=\qualextwo,height=\heightqualex]{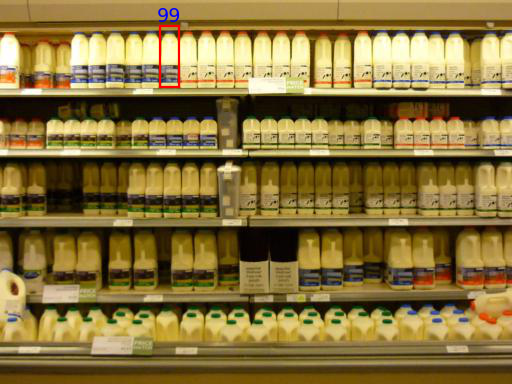} 
        \includegraphics[width=\qualextwo,height=\heightqualex]{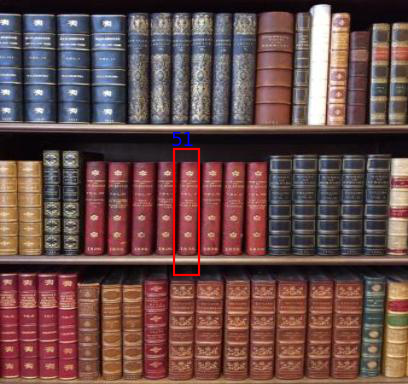}
    \includegraphics[width=\qualextwo,height=\heightqualex]{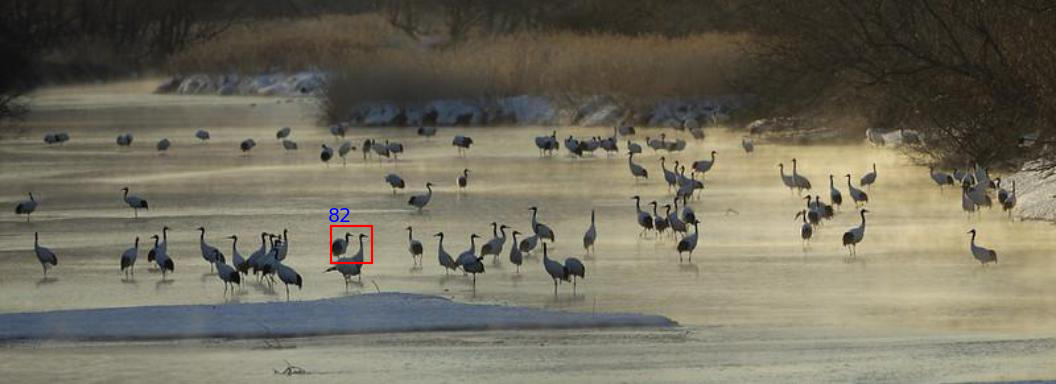} 
    \\    
\makebox[\qualextwo]{GT: 211 , Rep: 223}
\makebox[\qualextwo]{GT: 120, Rep: 99} 
\makebox[\qualextwo]{GT: 51, Rep: 51}
\makebox[\qualextwo]{GT: 84, Rep: 82} 
\\ \vspace{2ex}

    \includegraphics[width=\qualextwo,height=\heightqualex]{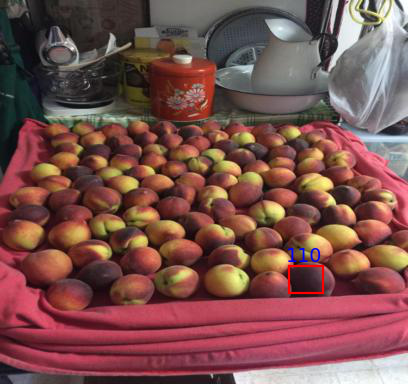}
    \includegraphics[width=\qualextwo,height=\heightqualex]{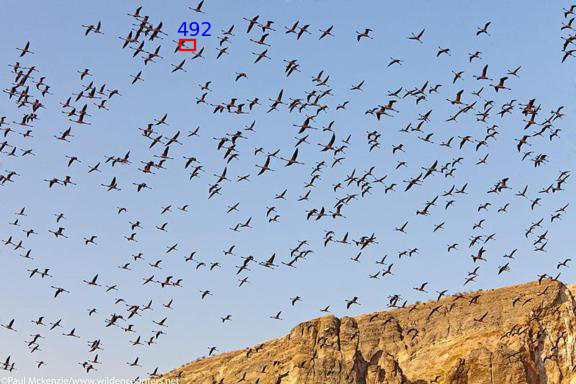} 
        \includegraphics[width=\qualextwo,height=\heightqualex]{Images1/3497_161_127_RPN.png}
    \includegraphics[width=\qualextwo,height=\heightqualex]{Images1/4969_108_91_RPN.png} 
    \\    
\makebox[\qualextwo]{GT: 108, Rep: 110}
\makebox[\qualextwo]{GT: 402, Rep: 492} 
\makebox[\qualextwo]{GT: 161, Rep: 127}
\makebox[\qualextwo]{GT: 108, Rep: 91} 
\\ \vspace{2ex}

    \includegraphics[width=\qualextwo,height=\heightqualex]{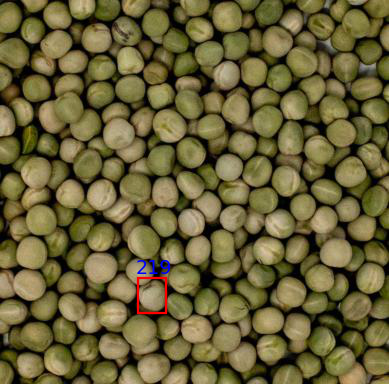}
    \includegraphics[width=\qualextwo,height=\heightqualex]{Images1/6249_110_109_RPN.png} 
        \includegraphics[width=\qualextwo,height=\heightqualex]{Images1/309_113_87_RPN.png}
    \includegraphics[width=\qualextwo,height=\heightqualex]{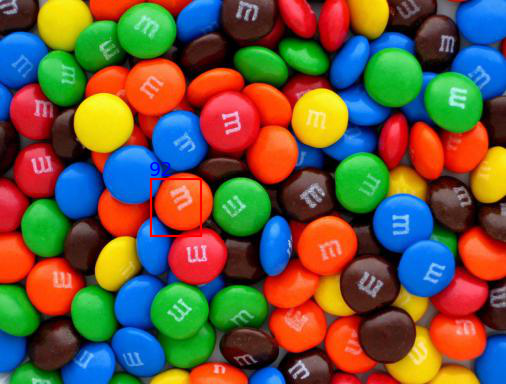} 
    \\    
\makebox[\qualextwo]{GT: 262, Rep: 219}
\makebox[\qualextwo]{GT: 110, Rep: 109} 
\makebox[\qualextwo]{GT: 113, Rep: 87}
\makebox[\qualextwo]{GT: 111, Rep: 92} 
\\ \vspace{2ex}

    \includegraphics[width=\qualextwo,height=\heightqualex]{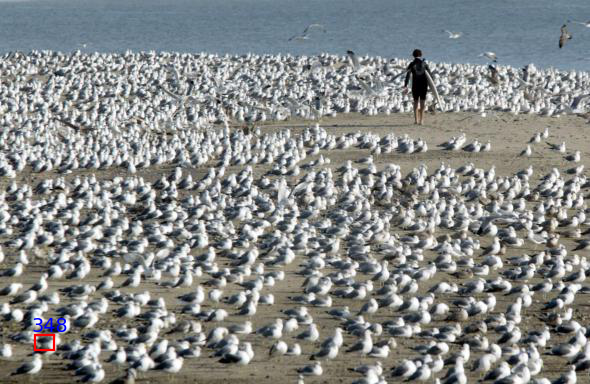}
    \includegraphics[width=\qualextwo,height=\heightqualex]{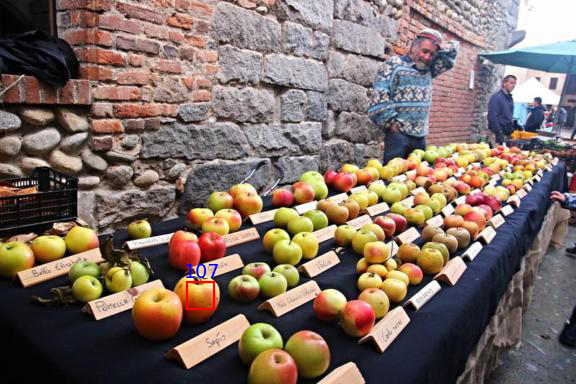}    
    \includegraphics[width=\qualextwo,height=\heightqualex]{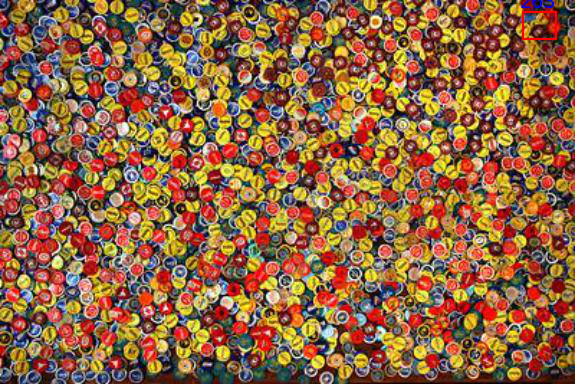} 
    \includegraphics[width=\qualextwo,height=\heightqualex]{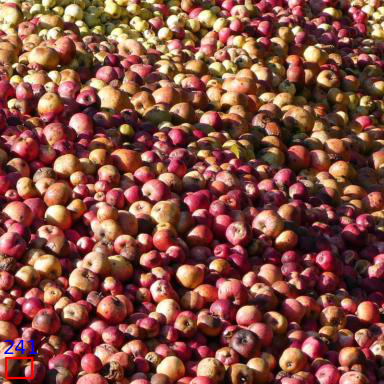} 
    \\    
\makebox[\qualextwo]{GT: 1092, Rep: 348}
\makebox[\qualextwo]{GT: 154, Rep: 107}
\makebox[\qualextwo]{GT: 1228, Rep: 263} 
\makebox[\qualextwo]{GT: 548, Rep: 241} 

  \vskip -0.1in
  \caption{{\bf Selected proposal (shown in red) and corresponding repetition score predicted by RepRPN}. The first three rows are success cases for RepRPN, and the predicted repetition score is close to the ground truth count. The last row shows failure cases, due to extreme scale variation and small object sizes.}
  \label{fig:QualitativeMore}
\end{figure*}

\section{Conclusions \label{sec:conclusion}}
In this paper, we tackled the task of Exemplar Free Class Agnostic Counting. We proposed RepRPN-Counter, the first exemplar free class agnostic counter capable of handling previously unseen categories at test time. Our two-stage counter consists of a novel region proposal network for finding the exemplars from repetitive object classes, and a density estimation network to estimate the density map corresponding to each exemplar. We also showed that our region proposal network can significantly improve the performance of the previous state-of-the-art class-agnostic visual counters.

\clearpage
%
%
\bibliographystyle{splncs04}
\bibliography{egbib}

\begin{thebibliography}{10}
\providecommand{\url}[1]{\texttt{#1}}
\providecommand{\urlprefix}{URL }
\providecommand{\doi}[1]{https://doi.org/#1}

\bibitem{m_Abousamra-etal-AAAI21}
Abousamra, S., Hoai, M., Samaras, D., Chen, C.: Localization in the crowd with
  topological constraints. In: AAAI (2021)

\bibitem{arteta2016detecting}
Arteta, C., Lempitsky, V., Noble, J.A., Zisserman, A.: Detecting overlapping
  instances in microscopy images using extremal region trees. Medical image
  analysis  \textbf{27},  3--16 (2016)

\bibitem{arteta2016counting}
Arteta, C., Lempitsky, V., Zisserman, A.: Counting in the wild. In: ECCV (2016)

\bibitem{babu2018divide}
Babu~Sam, D., Sajjan, N.N., Venkatesh~Babu, R., Srinivasan, M.: Divide and
  grow: Capturing huge diversity in crowd images with incrementally growing
  cnn. In: CVPR (2018)

\bibitem{bansal2018zero}
Bansal, A., Sikka, K., Sharma, G., Chellappa, R., Divakaran, A.: Zero-shot
  object detection. In: Proceedings of the European Conference on Computer
  Vision (ECCV). pp. 384--400 (2018)

\bibitem{cao2018scale}
Cao, X., Wang, Z., Zhao, Y., Su, F.: Scale aggregation network for accurate and
  efficient crowd counting. In: ECCV (2018)

\bibitem{detr}
Carion, N., Massa, F., Synnaeve, G., Usunier, N., Kirillov, A., Zagoruyko, S.:
  End-to-end object detection with transformers. In: European conference on
  computer vision. pp. 213--229. Springer (2020)

\bibitem{finn2017model}
Finn, C., Abbeel, P., Levine, S.: Model-agnostic meta-learning for fast
  adaptation of deep networks (2017)

\bibitem{He-etal-ICCV17}
He, K., Gkioxari, G., Doll{\'a}r, P., Girshick, R.: Mask {R-CNN}. In: ICCV
  (2017)

\bibitem{He-et-al-CVPR16}
He, K., Zhang, X., Ren, S., Sun, J.: Deep residual learning for image
  recognition. In: CVPR (2016)

\bibitem{Idrees_2013_CVPR}
Idrees, H., Saleemi, I., Seibert, C., Shah, M.: Multi-source multi-scale
  counting in extremely dense crowd images. In: CVPR (2013)

\bibitem{idrees2018composition}
Idrees, H., Tayyab, M., Athrey, K., Zhang, D., Al-Maadeed, S., Rajpoot, N.,
  Shah, M.: Composition loss for counting, density map estimation and
  localization in dense crowds. In: ECCV (2018)

\bibitem{khan2016deep}
Khan, A., Gould, S., Salzmann, M.: Deep convolutional neural networks for human
  embryonic cell counting. In: ECCV. Springer (2016)

\bibitem{kingma2014adam}
Kingma, D.P., Ba, J.: Adam: A method for stochastic optimization. arXiv
  preprint arXiv:1412.6980  (2014)

\bibitem{koch2015siamese}
Koch, G., Zemel, R., Salakhutdinov, R.: Siamese neural networks for one-shot
  image recognition. In: ICML deep learning workshop (2015)

\bibitem{lake2015human}
Lake, B.M., Salakhutdinov, R., Tenenbaum, J.B.: Human-level concept learning
  through probabilistic program induction. Science  \textbf{350}(6266),
  1332--1338 (2015)

\bibitem{li2018csrnet}
Li, Y., Zhang, X., Chen, D.: Csrnet: Dilated convolutional neural networks for
  understanding the highly congested scenes. In: CVPR (2018)

\bibitem{lin2017focal}
Lin, T.Y., Goyal, P., Girshick, R., He, K., Doll{\'a}r, P.: Focal loss for
  dense object detection. In: ICCV (2017)

\bibitem{lin2014microsoft}
Lin, T.Y., Maire, M., Belongie, S., Hays, J., Perona, P., Ramanan, D.,
  Doll{\'a}r, P., Zitnick, C.L.: Microsoft coco: Common objects in context. In:
  European conference on computer vision. pp. 740--755. Springer (2014)

\bibitem{liu2019context}
Liu, W., Salzmann, M., Fua, P.: Context-aware crowd counting. In: CVPR (2019)

\bibitem{Liu-etal-CVPR18}
Liu, X., van~de Weijer, J., Bagdanov, A.D.: Leveraging unlabeled data for crowd
  counting by learning to rank. In: CVPR (2018)

\bibitem{liu2018leveraging}
Liu, X., Van De~Weijer, J., Bagdanov, A.D.: Leveraging unlabeled data for crowd
  counting by learning to rank. In: CVPR (2018)

\bibitem{liu2020semi}
Liu, Y., Liu, L., Wang, P., Zhang, P., Lei, Y.: Semi-supervised crowd counting
  via self-training on surrogate tasks. In: European Conference on Computer
  Vision. pp. 242--259. Springer (2020)

\bibitem{lu2018class}
Lu, E., Xie, W., Zisserman, A.: Class-agnostic counting. In: ACCV (2018)

\bibitem{ma2019bayesian}
Ma, Z., Wei, X., Hong, X., Gong, Y.: Bayesian loss for crowd count estimation
  with point supervision. In: ICCV (2019)

\bibitem{mundhenk2016large}
Mundhenk, T.N., Konjevod, G., Sakla, W.A., Boakye, K.: A large contextual
  dataset for classification, detection and counting of cars with deep
  learning. In: ECCV (2016)

\bibitem{rahman2018zero}
Rahman, S., Khan, S., Porikli, F.: Zero-shot object detection: Learning to
  simultaneously recognize and localize novel concepts. In: Asian Conference on
  Computer Vision. pp. 547--563. Springer (2018)

\bibitem{ranjan2018iterative}
Ranjan, V., Le, H., Hoai, M.: Iterative crowd counting. In: ECCV (2018)

\bibitem{ranjan2019crowd}
Ranjan, V., Shah, M., Nguyen, M.H.: Crowd transformer network. arXiv preprint
  arXiv:1904.02774  (2019)

\bibitem{ranjan2021learning}
Ranjan, V., Sharma, U., Nguyen, T., Hoai, M.: Learning to count everything. In:
  Proceedings of the IEEE/CVF Conference on Computer Vision and Pattern
  Recognition. pp. 3394--3403 (2021)

\bibitem{m_Ranjan-etal-ACCV20}
Ranjan, V., Wang, B., Shah, M., Hoai, M.: Uncertainty estimation and sample
  selection for crowd counting. In: ACCV (2020)

\bibitem{ravi2016optimization}
Ravi, S., Larochelle, H.: Optimization as a model for few-shot learning  (2016)

\bibitem{ren2015faster}
Ren, S., He, K., Girshick, R., Sun, J.: Faster r-cnn: Towards real-time object
  detection with region proposal networks. In: NeurIPS (2015)

\bibitem{Ren-etal-NIPS15}
Ren, S., He, K., Girshick, R., Sun, J.: Faster {R-CNN}: Towards real-time
  object detection with region proposal networks. In: NeurIPS (2015)

\bibitem{sam2017switching}
Sam, D.B., Surya, S., Babu, R.V.: Switching convolutional neural network for
  crowd counting. In: CVPR (2017)

\bibitem{santoro2016one}
Santoro, A., Bartunov, S., Botvinick, M., Wierstra, D., Lillicrap, T.: One-shot
  learning with memory-augmented neural networks  (2016)

\bibitem{shi2019revisiting}
Shi, M., Yang, Z., Xu, C., Chen, Q.: Revisiting perspective information for
  efficient crowd counting. In: CVPR (2019)

\bibitem{sindagi2020jhu}
Sindagi, V.A., Yasarla, R., Patel, V.M.: Jhu-crowd++: Large-scale crowd
  counting dataset and a benchmark method. arXiv preprint arXiv:2004.03597
  (2020)

\bibitem{song2021rethinking}
Song, Q., Wang, C., Jiang, Z., Wang, Y., Tai, Y., Wang, C., Li, J., Huang, F.,
  Wu, Y.: Rethinking counting and localization in crowds: A purely point-based
  framework. In: Proceedings of the IEEE/CVF International Conference on
  Computer Vision. pp. 3365--3374 (2021)

\bibitem{vaswani2017attention}
Vaswani, A., Shazeer, N., Parmar, N., Uszkoreit, J., Jones, L., Gomez, A.N.,
  Kaiser, {\L}., Polosukhin, I.: Attention is all you need. In: Advances in
  neural information processing systems. pp. 5998--6008 (2017)

\bibitem{wan2019adaptive}
Wan, J., Chan, A.: Adaptive density map generation for crowd counting. In:
  Proceedings of the IEEE International Conference on Computer Vision. pp.
  1130--1139 (2019)

\bibitem{wan2021generalized}
Wan, J., Liu, Z., Chan, A.B.: A generalized loss function for crowd counting
  and localization. In: Proceedings of the IEEE/CVF Conference on Computer
  Vision and Pattern Recognition. pp. 1974--1983 (2021)

\bibitem{wang2021uniformity}
Wang, C., Song, Q., Zhang, B., Wang, Y., Tai, Y., Hu, X., Wang, C., Li, J., Ma,
  J., Wu, Y.: Uniformity in heterogeneity: Diving deep into count interval
  partition for crowd counting. In: Proceedings of the IEEE/CVF International
  Conference on Computer Vision. pp. 3234--3242 (2021)

\bibitem{wang2020nwpu}
Wang, Q., Gao, J., Lin, W., Li, X.: Nwpu-crowd: A large-scale benchmark for
  crowd counting. arXiv preprint arXiv:2001.03360  (2020)

\bibitem{wang2019learning}
Wang, Q., Gao, J., Lin, W., Yuan, Y.: Learning from synthetic data for crowd
  counting in the wild. In: CVPR (2019)

\bibitem{wu2019detectron2}
Wu, Y., Kirillov, A., Massa, F., Lo, W.Y., Girshick, R.: Detectron2 (2019)

\bibitem{xie2018microscopy}
Xie, W., Noble, J.A., Zisserman, A.: Microscopy cell counting and detection
  with fully convolutional regression networks. Computer methods in
  biomechanics and biomedical engineering: Imaging \& Visualization
  \textbf{6}(3),  283--292 (2018)

\bibitem{zhang2019attentional}
Zhang, A., Yue, L., Shen, J., Zhu, F., Zhen, X., Cao, X., Shao, L.: Attentional
  neural fields for crowd counting. In: ICCV (2019)

\bibitem{zhang2016single}
Zhang, Y., Zhou, D., Chen, S., Gao, S., Ma, Y.: Single-image crowd counting via
  multi-column convolutional neural network. In: CVPR (2016)

\bibitem{zhu2019zero}
Zhu, P., Wang, H., Saligrama, V.: Zero shot detection. IEEE Transactions on
  Circuits and Systems for Video Technology  \textbf{30}(4),  998--1010 (2019)

\end{thebibliography}
\end{document}